\documentclass[runningheads]{llncs}
\usepackage[square,numbers]{natbib}
\usepackage[T1]{fontenc}
\usepackage{subcaption}
\usepackage{amsmath}
\usepackage{amssymb}
\usepackage{graphicx}
\usepackage{float}
\begin{document}
\title{Post-hoc and manifold explanations analysis of facial expression - psychological traits data based on deep learning}
%
%
\author{Yang Xiao\inst{1}}
\authorrunning{X. Yang}
%
\institute{Nankai University, Tianjin, China, \email{2013241@mail.nankai.edu.cn}}
\maketitle              
\begin{abstract}
The complex information processing system of humans generates a lot of objective and subjective evaluations, making the exploration of human cognitive products of great cutting-edge theoretical value. In recent years, deep learning technologies, which are inspired by biological brain mechanisms, have made significant strides in the application of psychological or cognitive scientific research, particularly in the memorization and recognition of facial data. This paper investigates through experimental research how neural networks process and store facial expression data and associate these data with a range of psychological attributes produced by humans. Researchers utilized deep learning model VGG16, demonstrating that neural networks can learn and reproduce key features of facial data, thereby storing image memories. Moreover, the experimental results reveal the potential of deep learning models in understanding human emotions and cognitive processes and establish a manifold visualization explanation of cognitive products or psychological attributes from a non-Euclidean space perspective, offering new insights into enhancing the explainability of AI. This study not only advances the application of AI technology in the field of psychology but also provides a new psychological theoretical understanding the information processing of the AI. The code is available in https://github.com/NKUShaw/Psychoinformatics.
\end{abstract}
\section{Introduction}
Recent years, deep learning technology has shown excellent performance in multiple fields and tasks, including weather prediction~\cite{bi_accurate_2023}, traffic prediction~\cite{nguyen_deep_2018}, financial prediction~\cite{heaton_deep_2017}, and natural language processing~\cite{vaswani_attention_2017}. With the development of computational power and the improvement of dataset quality, the utility of models has reached an unprecedented level. This progress is not only reflected in the models' ability to handle complex tasks but also in their capacity to understand and generate the diverse array of data provided by humans—including language and images—more accurately and efficiently, as well as to carry out multitask learning~\cite{nguyen2019multi}. The enhancement of this utility has made the application of deep learning technologies possible in various fields such as healthcare, automated manufacturing, intelligent security, educational technology, and environmental monitoring. Specifically, data is dimensionally reduced to more abstract concepts or representations at various levels within artificial neural networks, which is one of the core characteristics of how deep learning works~\cite{lecun2015deep}. Neural networks, through their multi-layered structures, are able to gradually extract and transform the features of input data, thus achieving the transformation from the initial raw data to the final high-level abstract representation. In layman's terms, these deep learning models are black-box models, whose internal abstracted data processing is opaque to the user~\cite{zhang2018interpretable}.

The rapid advancement of this technology and its black-box nature have raised an important question: how can we understand and explain the complex decisions made by these models in their decision-making process? This issue becomes even more pressing in the fields of psychology and cognitive science, which are dedicated to understanding the fundamental mechanisms of human thought and behavior. The lack of decision transparency in deep learning models not only challenges the boundaries of the technical field but also touches on the fundamental issue of how to use these technologies to simulate, understand, and even enhance human psychological processes. Therefore, exploring the decision-making processes and data handling mechanisms within these models in conjunction with psychological research can not only help us improve the interpretability and transparency of AI but may also provide new insights into understanding the human brain.

Take facial data as an example: the face is the front part of a person or other living organism, especially the front part of the head, which includes features such as eyes, nose, and mouth. In humans, the face is not only a significant marker of identity but also a key tool for emotional expression, social interaction, and non-verbal communication. Humans are capable of recognizing and explaining subtle facial expressions to understand others' emotional states, intentions, and personality traits. This information processing mirrors the process of data handling in neural networks, where a network can abstract a whole face into unique features (such as the relative position of eyes, nose, and mouth, facial contours, muscle changes, etc.). The activation functions of convolutional neural networks and the operating mechanisms of convolutional layers are highly similar to the Hodgkin-Huxley model~\cite{hodgkin_quantitative_1952} and the visual cortex~\cite{lin_four_2021,furlan_global_2016,bullier_integrated_2001,zeki_area_2015}, which means that neural networks can serve as a simplified mathematical model to simulate the information processing mechanisms of the biological brain.

Behaviorists used conditioned reflex experiments, like Pavlov's dog\cite{pavlov2010conditioned} 
and Skinner's box\cite{skinner2019behavior}
, to explore the relationships between behavior and environmental stimuli, thus explainaing the "psychological black-box." Similarly, in the field of Explainable AI, scholars have developed serveral post-hoc explainable algorithms like LIME (Local Interpretable Model-agnostic Explanations)\cite{ribeiro2016should}
, SHAP (SHapley Additive exPlanations)\cite{lundberg2017unified}
, and DeepLIFT (Layer-wise Relevance Propagation) \cite{shrikumar2017learning}to try to explain how input data affects a model's predictive output. These methods advocate for objective observation under controlled laboratory conditions, and seek scientific rigor through repeated experiments and emphasis on the measurement of external behaviors. 

Due to human sociality and interactivity, facial expression data naturally contains a wealth of information implicit psychological or social attributes\cite{ekman1979facial, kanade2000comprehensive, morris1996differential, duncan2015face}. Neural networks have achieved many outstanding results in training facial expression data, such as political affiliation prediction\cite{berg2022can}, social judgments prediction\cite{keles2021cautionary}, ideology prediction\cite{rasmussen2023using}, BIG FIVE prediction\cite{kachur2020assessing,nilugonda2021big}, and sexual orientation\cite{wang2018deep, leuner2019replication}. However, they do not delve into the underlying patterns of psychological variables or the data acquired internally by neural networks. The success in engineering applications cannot conceal their theoretical scarcity:
\begin{itemize}
    \item Lack of Transparency, the internal workings of the layers and the complex transformations they apply to the data are not transparent. 
    \item Lack of Theoretical Framework, neural networks might not understand the reason why certain patterns are predictive.
    \item Surface-Level Analysis, neural networks might not learn these deeper and causative laws that genuinely underlie those traits.
    \item Complexity of Psychological Attributes: Psychological variables are abstract concepts that encompass a wide range of behaviors and internal states. A neural network's prediction is typically a reductionist view of these complex constructs.
    \item Interconnectedness of Psychological Attributes: In psychology, variables often influence one another in a highly interconnected and dynamic system, which a neural network might not fully capture or explain due to its static nature.
\end{itemize}. 

Psychological research involves a complex and diverse array of variables, from human behavior to cognitive processes, and emotional states. The interactions among these variables are typically multidimensional and intricate, and cannot be simply represented or measured through traditional geometric methods. Although they may manifest in a high-dimensional space, are in fact likely distributed on a low-dimensional manifold with complex geometric structures. Psychological attributes, such as personality, emotional states, cognitive styles, and so on, are often multifaceted and complex, and their expressions can be very rich and nuanced\cite{liu_deep_2016}. The criterion of good explainable: the feature attributions need to be aligned with the tangent space of the data manifold\cite{bordt2023manifold}. As a consequence, they suggest that explanation algorithms should actively strive to align their explanations with the data manifold.

We use the convolutional neural network VGG16\cite{simonyan_very_2014} 
to simulate the information processing mechanisms of the biological brain, surpassing human predictive power and geometric feature extraction methods across 40 relative psychological variables. We use post-hoc explainable algorithms grad-CAM\cite{selvaraju_grad-cam_2017} 
to explain the neural network's decision outputs concerning psychological variables. Most importantly, we use manifold learning algorithms, which study complex structural relationships, to reveal the intrinsic relationships acquired by the neural network between facial expression image data and psychological variable data through internal explanations.

Our contribution is following:
\begin{itemize}
    \item Get \textbf{SOTA} result in Wilma Bainbridge 10k US Adult Faces Database\cite{bainbridge_intrinsic_2013} through VGG16.
    \item Attribute the connection between psychological variables and facial features through grad-CAM.
    \item Explain the internal patterns of psychological attributes predicted by neural networks through explainable algorithms like t-SNE and UMAP, providing a new avenue of research for AI in Psychology.
\end{itemize}

\section{Related Works}
    \subsection{Interdisciplinary background}
    The Hebbian learning rule\cite{hebb2005organization} 
    is widely considered a fundamental principle in the study of artificial neural networks, stating, "When neuron A sufficiently frequently partakes in firing neuron B, such that it increases the probability of neuron B's activation, the synaptic connection between them is strengthened." This rule is also vividly described as "cells that fire together, wire together." The core of Hebbian learning is synaptic plasticity, which means that neural networks learn by adjusting the strengths of the synapses between neurons. Although Hebb himself did not directly work in the field of artificial neural networks, his theory provided later researchers with a method to simulate the brain's learning process. In artificial neural networks, the Hebbian learning rule is used to model the process of adjusting weights between neurons.

    Frank Rosenblatt's initial proposal of the perceptron model\cite{rosenblatt1958perceptron} 
    spurred the development of the Multi-Layer Perceptron (MLP). The physiological research by Alan Hodgkin and Andrew Huxley, known as the Hodgkin-Huxley model, provided biological validation for the theoretical basis of nonlinear activation functions, inspiring their proposal and application\cite{rumelhart1986learning, hahnloser2000digital, bridle1990probabilistic, nair2010rectified, maas2013rectifier}. 
    This enabled artificial neural networks to acquire nonlinear modeling capabilities. Such networks, composed of multiple layers of neurons, utilize these nonlinear capabilities to learn complex patterns and features from input data. Consequently, MLP has become a fundamental basis for deep learning, thereby supporting the development of modern, complex multilayer neural network architectures. This series of developments demonstrates that the evolution of artificial neural networks from simple models to complex network structures has been driven by contributions from multiple interdisciplinary fields, including biology, psychology, mathematics, and computer science\cite{lecun2015deep}.
    \subsection{Post-hoc explanation}
    Post-hoc explanations in machine learning refer to techniques and methods applied after a model has been trained to claim how it makes decisions or predictions\cite{slack2021reliable}. 

    \textbf{LIME}(Local Interpretable Model-agnostic Explanations) provides local explanations for individual predictions by approximating the model locally with an interpretable model\cite{ribeiro2016should}. 
    \textbf{SHAP}(Shapley Additive explanations) utilizes game theory to determine the contribution of each feature to the prediction by considering all possible combinations of features\cite{lundberg2017unified}.
    \textbf{Grad-CAM}(Gradient-weighted Class Activation Mapping) generates visual explanations for models involved in image-related tasks, highlighting the important regions in the image for predicting the model’s output\cite{selvaraju_grad-cam_2017}.
    \textbf{Counterfactual Explanations} focuses on explaining model decisions by showing how slight changes to the input features could lead to different predictions\cite{guidotti2022counterfactual}.
    \textbf{Integrated Gradients} attributes the prediction of a deep network to its input features based on the gradients of the output with respect to the input\cite{qi2019visualizing}.These explanations are crucial for understanding machine learning or deep learning models those frequently classified as black-box\cite{zhao2021causal}. 

    Therefore, the growing reliance on machine learning in critical areas such as healthcare, finance, and law enforcement has heightened the importance of transparency and accountability, making post-hoc explanations an active area of research\cite{hanif2021survey}.
    \subsection{Manifold Learning}
    Manifold Learning is a machine learning approach to exploring and understanding the structure of high-dimensional data\cite{echeverria_manifold_2008}. 
    It posits that high-dimensional data are distributed within a lower-dimensional space, forming a manifold embedded within the high-dimensional space. The primary goal of manifold learning is to discover the intrinsic low-dimensional structures underlying these high-dimensional data points, facilitating easier analysis and visualization. Although many real-world datasets may exist in very high dimensions, such as images, sound, bioinformatics, psychological or text data, the essential dimensions of variation are relatively few\cite{cheng_sagman_2024}. These essential changes can be considered a kind of low-dimensional manifold structure\cite{xiao_one_2015}. 

    \textbf{LLE}(Locally Linear Embedding) reduces dimensionality while trying to preserve the geometric features of the input data at a local level\cite{roweis2000nonlinear}.
    \textbf{Isomap} extends Multidimensional Scaling (MDS)\cite{shepard1962analysis1, shepard1962analysis2} by using geodesic distances on the manifold instead of the Euclidean distances in the input space\cite{tenenbaum2000global}. 
    \textbf{t-SNE}(t-Distributed Stochastic Neighbor Embedding) converts affinities of data points to probabilities and tries to minimize the Kullback-Leibler divergence between the joint probabilities of the low-dimensional embedding and the high-dimensional data \cite{van2008visualizing}.
    \textbf{Diffusion Maps} provide a framework for extracting significant features from data by viewing data as a probability distribution over a graph. It uses a diffusion process to reveal the geometric structure of the data at different scales, effectively highlighting the connections within the data\cite{coifman2006diffusion}.
    \textbf{UMAP}(Uniform Manifold Approximation and Projection) is a modern manifold learning technique for dimension reduction that is both scalable and effective at preserving both local and global data structure. This technique works similarly to t-SNE but is often faster and scales better to larger datasets\cite{mcinnes2018umap}.
\section{Preliminary}
The most commonly used post hoc explanation methods for computing future importance are based on gradient computations. Let $f:x\in\mathbb{R}^{d}\to y$ be any arbitrary classifier, the gradient $\nabla f(x,y)\in\mathbb{R}^{T}$ of the model $f$ is given as:
\begin{equation}\label{Gradient}
    \nabla f(x,y) = \frac{\partial\mathcal{L}(x,y;\omega)}{\partial\omega}
\end{equation}
where $\mathcal{L}$ is the loss function of cross-entropy for classification in our experiments and $\omega$ is the parameters of $f$.

\section{Explanation}
\subsection{Grad-CAM}
Grad-CAM uses the gradients of any target concept flowing into the final convolutional layer to produce a coarse localization map highlighting the important regions in the image for predicting the concept. Mathematically, it can be expressed as:
\begin{equation}\label{Grad-CAM}
    \phi_\text{Grad-CAM} = Activation(\sum_{k}\alpha^{c}_{k}A^k)
\end{equation}
where $A^k$ are the feature maps of the convolutional layer, $c$ is a class output, and $\alpha^c_k$ are the weights calculated as the global average of the gradients:
\begin{equation}
    \alpha_{k}^{c}=\frac{1}{Z}\sum_{i}\sum_{j}\frac{\partial f(x)^c}{\partial A^k_{ij}}
\end{equation}

\subsection{t-SNE}
Now, deep learning model predicted a series of psychological data:
\begin{equation}
    \mathbf{M}_{\text{predicted}}=f(X;\omega)
\end{equation}
we separate psychological labels $T$ from $\mathbf{M}_{\text{predicted}}$ and $\mathbf{M}$ represent the result of $\mathbf{M}_{\text{predicted}}$.

Let $\mathbf{M}={\mathbf{m_1},\mathbf{m_2},...,\mathbf{m_n}}$ represent the high-dimensional data points where each $\mathbf{m_i}\in \mathbb{R}^d$.
The pairwise similarities in the high-dimensional space are calculated using Gaussian distributions centered at each data point. The probability $p_{j|i}$ that point $\mathbf{m_j}$ is neighbor of point $\mathbf{m_i}$(conditional probability) is given by:

\begin{equation}
    p_{j|i} = \frac{\exp\left(-\|\mathbf{m}_i - \mathbf{m}_j\|^2 / 2\sigma_i^2\right)}{\sum_{k \neq i} \exp\left(-\|\mathbf{m}_i - \mathbf{m}_k\|^2 / 2\sigma_i^2\right)}
\end{equation}
where $\sigma_i$ is the variance of the Gaussian that is centered on point $\mathbf{m_j}$. This variance can be adapted for each point to accommodate different densities (perplexity setting). We can obtain joint probabilities by:
\begin{equation}
    \mathbf{P}=\frac{p_{j|i}+p_{i|j}}{2n}
\end{equation}

In the low-dimensional space, similarities are calculated using a Student's t-distribution (which has heavier tails than the Gaussian used in the high-dimensional space) to allow for a better modeling of distances between far-apart points. The probability $q_{ij}$ is given by:
\begin{equation}
    q_{ij}=\frac{(1+\lVert t_i - t_j \rVert^2)^{-1}}{\sum_{k\neq l}(1+\lVert t_k - t_l \rVert^2)^{-1})}
\end{equation}

Finally, we can get the low-dimensional embedding:
\begin{equation}
    \text{minimize}(KL(P\|Q))=\sum_{i}\sum_{j}p_{ij}\log\frac{p_{ij}}{q_{ij}}
\end{equation}
where $KL$ is Kullback-Leibler divergence.
\subsection{UMAP}
For high-dimensional distances,
\begin{equation}
    d_{i,j}^{\text{high}} = \| \mathbf{m}_i - \mathbf{m}_j \| 
\end{equation}
For low-dimensional distance after UMAP reduction,
\begin{equation}
    d_{i,j}^{\text{low}} = \| \mathbf{t}_i - \mathbf{t}_j \|     
\end{equation}
UMAP constructs a high-dimensional graph representation of the data and then optimizes a low-dimensional graph to be as structurally similar as possible. The objective function is somewhat similar to t-SNE but focuses on preserving the topological structure:
\begin{equation}
    \text{minimize}\sum_{i,j}(d_{i,j}^{high}-d_{i,j}^{low})^2
\end{equation}
\section{Experiments}
In this section, we first describe the  10k US Adult Face Database\cite{bainbridge_intrinsic_2013} used in our experiments. Next, we introduce how we design experiments to achieve optimal results on 40 psychological attributes. Subsequently, we explain how to post-hoc explain the feature maps of convolutional neural networks and why a specific convolutional layer is chosen. Finally, we introduce manifold learning algorithms to model the predicted set of psychological attributes to provide intrinsic explanations.
\subsection{Dataset}
We use 2222 labeled images of Wilma Bainbridge 10k US Adult Faces Database. During training, divide into training set : validation set : test set=64:16:20.

The psychological attribute labels are scored by participants recruited by the dataset creator\cite{bainbridge_intrinsic_2013}. Due to the influence of factors such as the race, age, and gender of the subjects themselves, as well as the race, age, and gender of the face images in the dataset, psychological attributes may be affected. In previous studies, researchers randomly divided the subjects into two groups, calculated the Pearson correlation based on the scoring results of the two groups, and repeated 50 times to take the average result as the predictive power of human psychological attributes\cite{song_social_2020}. Therefore, the closer the predictive power of the neural network is to 1.0, the better it can exclude the influence of additional variables.

The dataset consists of  20 pairs of psychological attributes. Each attribute is rated on a scale of 1-9 (1 means not at all, 9 means extremely). The 20 pairs of social traits are: (attractive, unattractive), (happy, unhappy), (friendly, unfriendly), (sociable, introverted), (kind, mean), (caring, cold), (calm, aggressive), (trustworthy, untrustworthy), (responsible, irresponsible),(confident, uncertain), (humble, egotistical),(emotionally stable, emotionally unstable), (normal, weird), (intelligent, unintelligent), (interesting, boring), (emotional, unemotional), (memorable, forgettable), (typical, atypical), (familiar, unfamiliar) and (common, uncommon).

\begin{figure*}[ht]
    \centering
    \begin{subfigure}{0.45\textwidth}
        \includegraphics[width=1.0\linewidth]{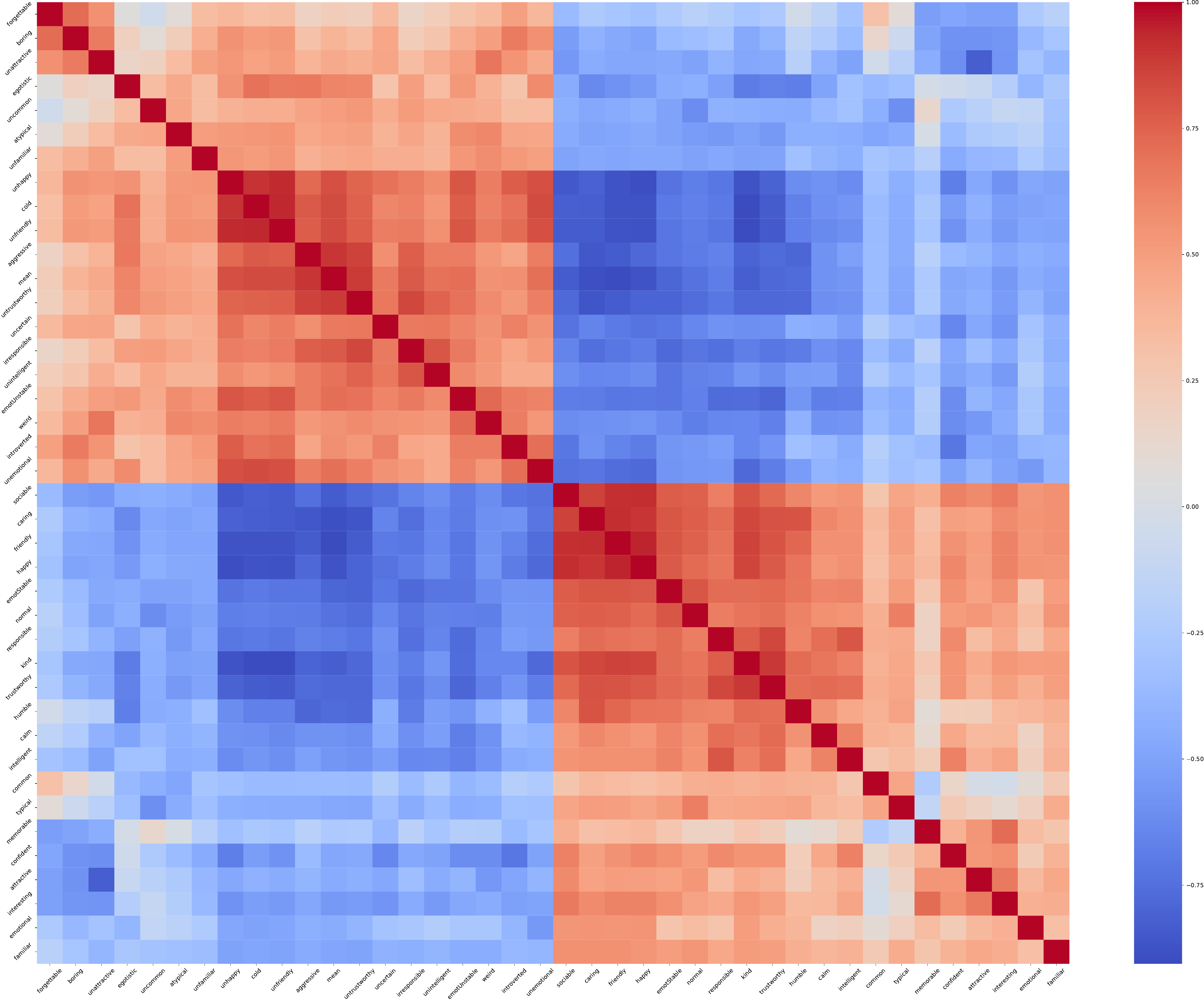}
        \caption{Heatmap for human}\label{fig:heatmap_human}
    \end{subfigure}
    \hspace{1cm}
    \begin{subfigure}{0.45\textwidth}
        \includegraphics[width=1.0\linewidth]{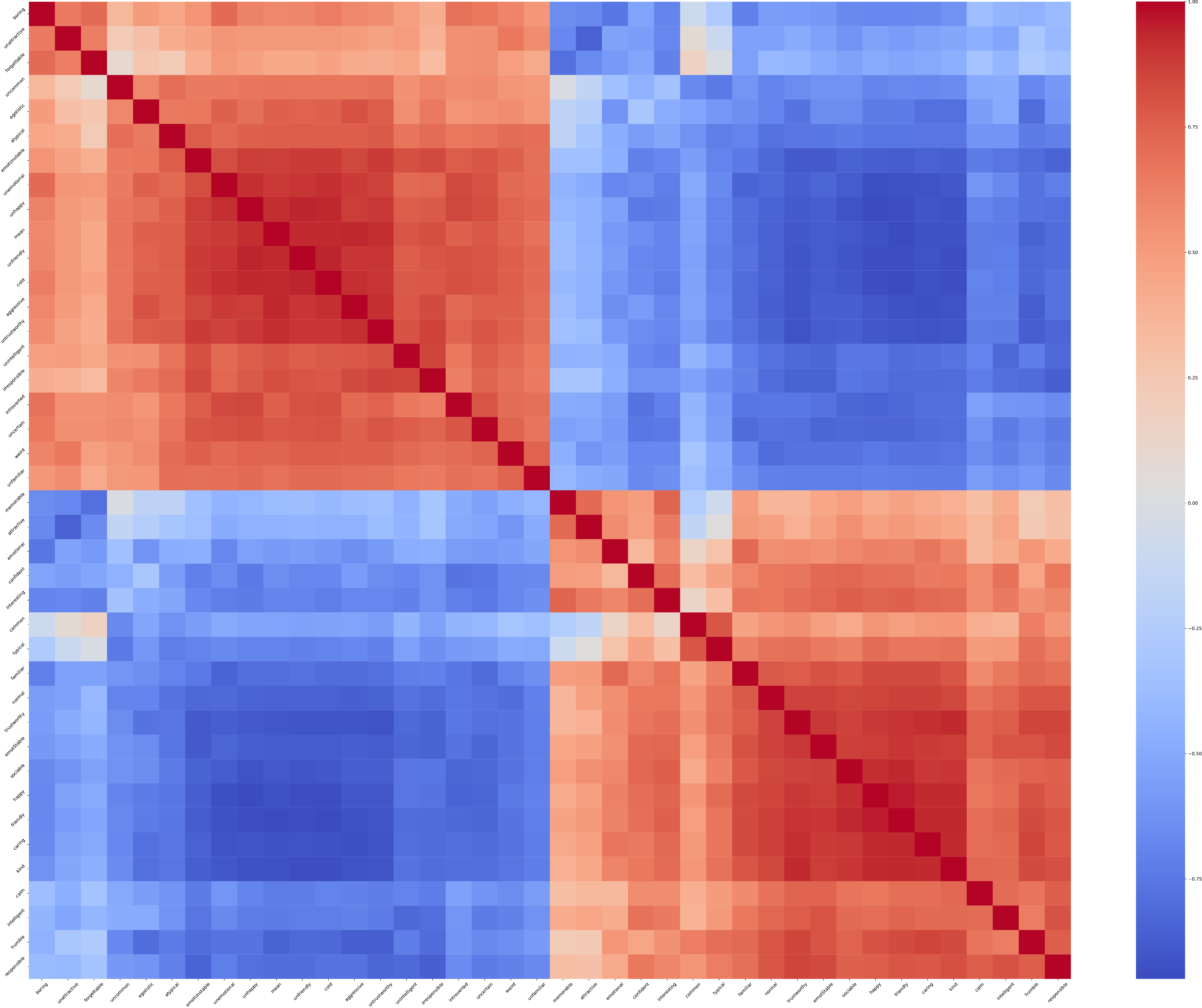}
        \caption{Heatmap for neural networks}\label{fig:heatmap_neuralnetwork}
    \end{subfigure}
    \caption{Heatmaps for human and neural network}\label{fig:heatmap}
\end{figure*}

Obviously, many psychological attributes are related to each other. We firstly calculates the correlation matrix for the dataset using the Pearson method. Through the linkage function, we do hierarchical/agglomerative cluster on the correlation matrix using the Ward variance minimization algorithm. Then, we constructs a dendrogram that displays the arrangements of the clusters produced by the corresponding linkage. We use the dendrogram to  determine the new order of rows (and columns, since it's a square matrix) that groups similar attributes together based on the clustering. Finally, we visulize the new heatmaps(see fig.\ref{fig:heatmap}). Compared to human annotated values, neural network predicts better clustering of psychological attributes and are more explainable.

\subsection{Training}
Considering the commonality between the local receptive field of convolutional layer and the biological visual cortex of the brain, we choose convoluntional neural networks to train this dataset. Song et al. obtained the conclusion that VGG16 has SOTA predictive power \cite{song2017learning}
through multiple comparisons such as VGG16, 
AlexNet\cite{krizhevsky_imagenet_2012}, and 
Inception from Google\cite{szegedy2015going}.

Local Receptive Fields have a limited field of view, only capturing a small portion of the input. Within a single layer, they may not capture patterns that span a broader area, and they might not be robust to more complex transformations such as scaling, rotation, or deformation\cite{sabour_dynamic_2017}. Furthermore, convolutional layers can be relatively insensitive to the exact positioning of features. To address this issue, we employ random horizontal flipping and random rotations within a range during training. By augmenting the dataset, we enhance the network's invariance to changes in position and orientation, aiding the model in learning more generalized features instead of relying solely on specific locations or directions.

According to experimental comparison, the optimal hyperparameter under our framework is: batch size=24 learning rate=5e-4. Compared with human predictive power and manual feature extraction methods\cite{song_social_2020}, neural networks have achieved all-round surpassing (see table \ref{tab:results}).

\begin{table}[ht]
    \centering
    \begin{tabular}{lcccc}
    \hline
    Attributes & Human & Geometric & VGG16 & Our Method \\ \hline
    Happy & 0.84 & 0.86 & 0.84 & \textbf{0.91} \\
    Unhappy & 0.75 & 0.81 & 0.8 & \textbf{0.87} \\
    Friendly & 0.78 & 0.83 & 0.82 & \textbf{0.88} \\
    Unfriendly & 0.72 & 0.8 & 0.79 & \textbf{0.83} \\
    \hline
    \end{tabular}
    \caption{A part of Predictive Power (Pearson Correlation Coefficients)}\label{tab:result}
\end{table}
\subsection{Post-hoc explanation}
\begin{figure}
    \centering
    \begin{minipage}{0.35\textwidth}
        \centering
        \includegraphics[width=\linewidth]{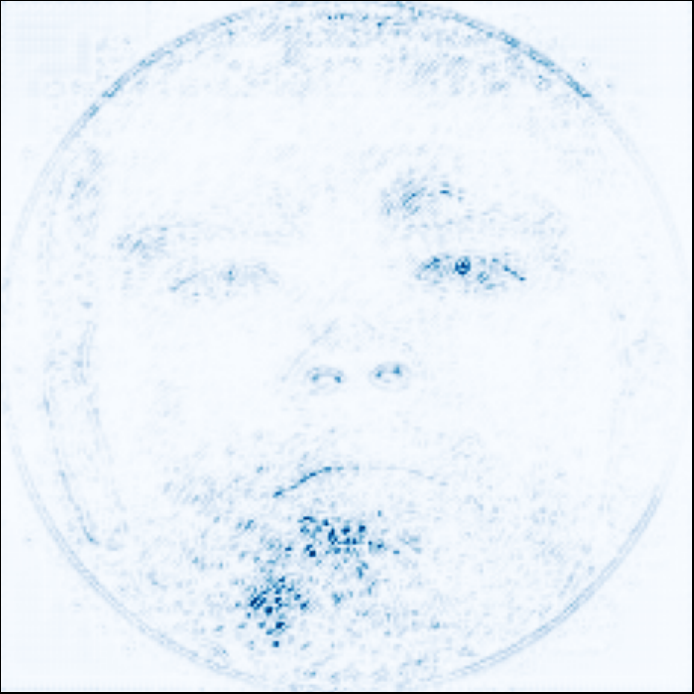}
    \end{minipage}
    \begin{minipage}{0.35\textwidth}
        \centering
        \includegraphics[width=\linewidth]{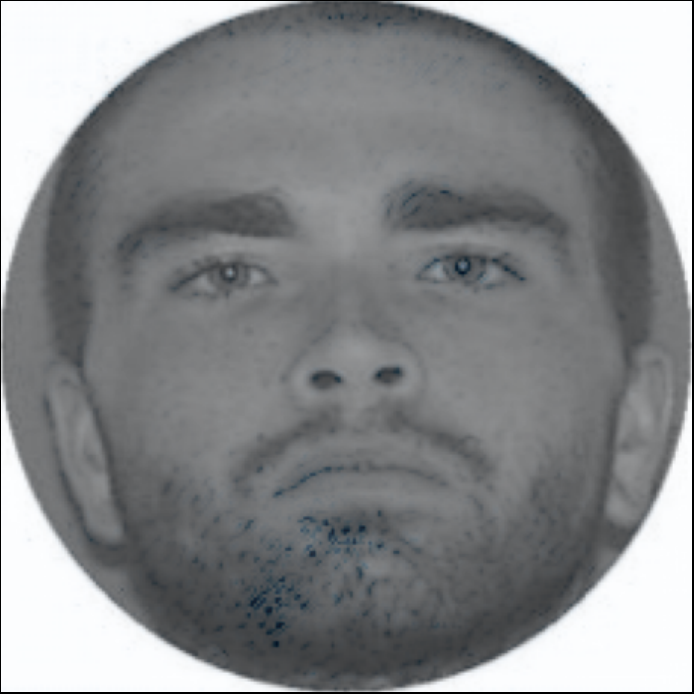}
    \end{minipage}
    \centering
    \begin{minipage}{0.35\textwidth}
        \centering
        \includegraphics[width=\linewidth]{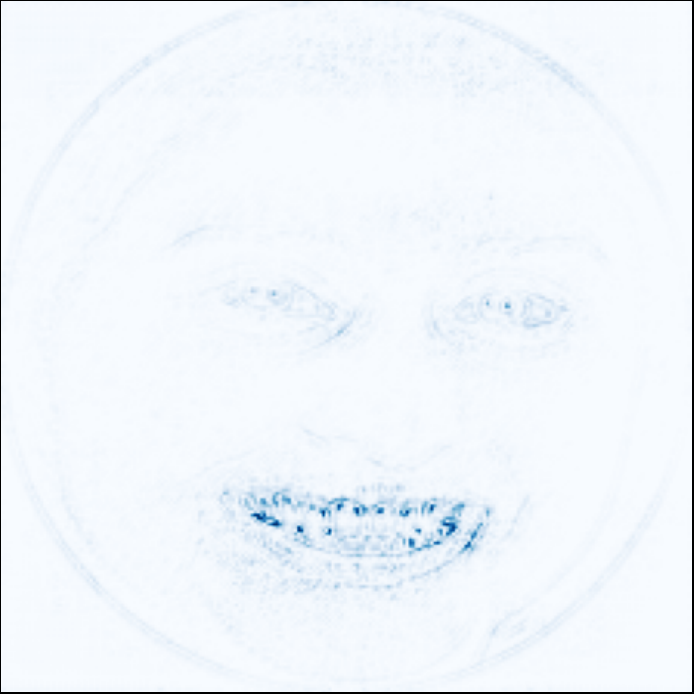}
    \end{minipage}
    \begin{minipage}{0.35\textwidth}
        \centering
        \includegraphics[width=\linewidth]{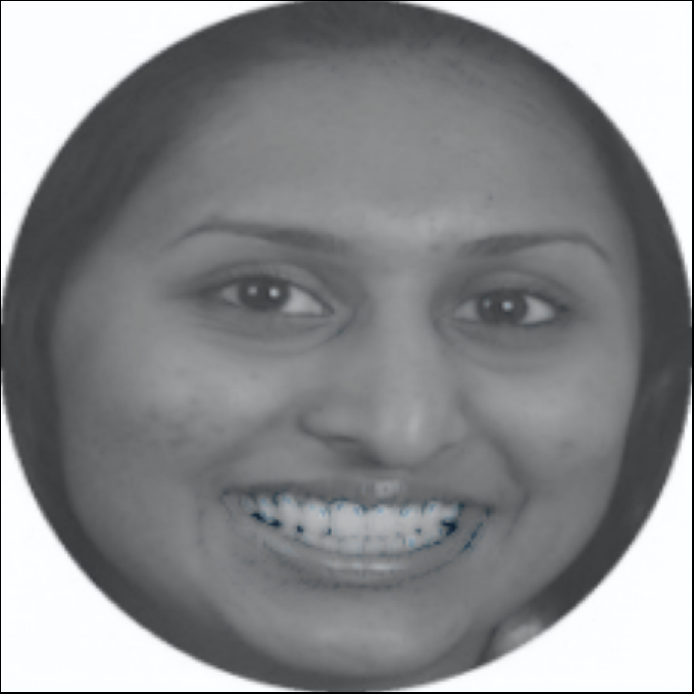}
    \end{minipage}
    \caption{Target: happy}\label{fig:post-hoc}
\end{figure}

According to formula \ref{Grad-CAM}, we calculated the convolutional layers with the highest weight values in the feature map and visualized them using Grad-CAM. In fig.\ref{fig:post-hoc}, these explanations suggests the model is focusing on the central features of the face—likely the eyes and mouth, which are crucial for explainaing happiness. The blended image overlays this heatmap on the original image, providing a more intuitive understanding of the heatmap's indications. The results clearly demonstrate that convolutional neural networks can effectively extract semantically meaningful facial features, associating psychological attributes with facial features.

\subsection{Manifold learning}
\begin{figure*}
    \centering
    \begin{subfigure}{0.4\textwidth}
        \includegraphics[width=1.0\linewidth]{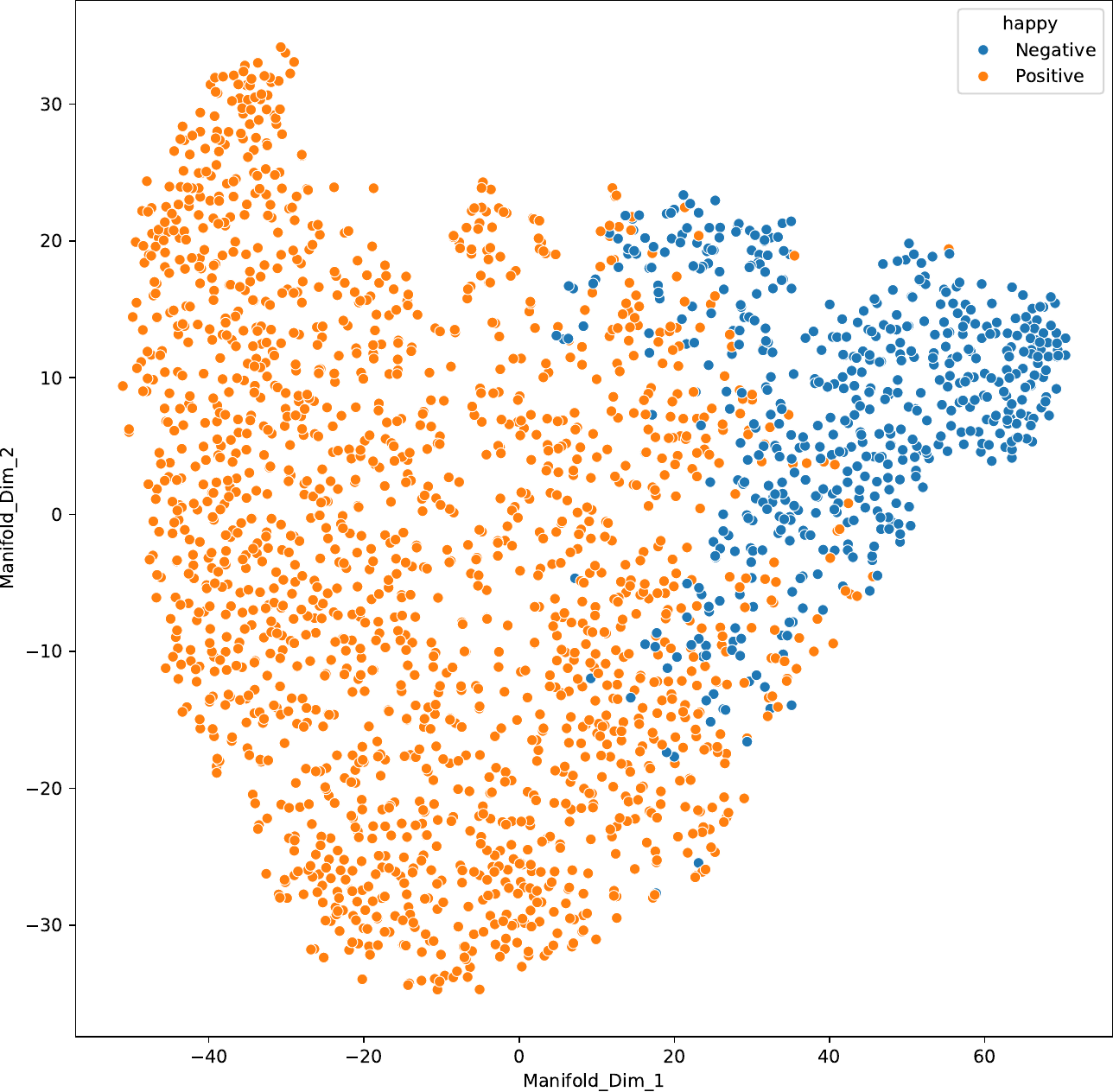}
        \caption{t-SNE for happy}
    \end{subfigure}
    \begin{subfigure}{0.4\textwidth}
        \includegraphics[width=1.0\linewidth]{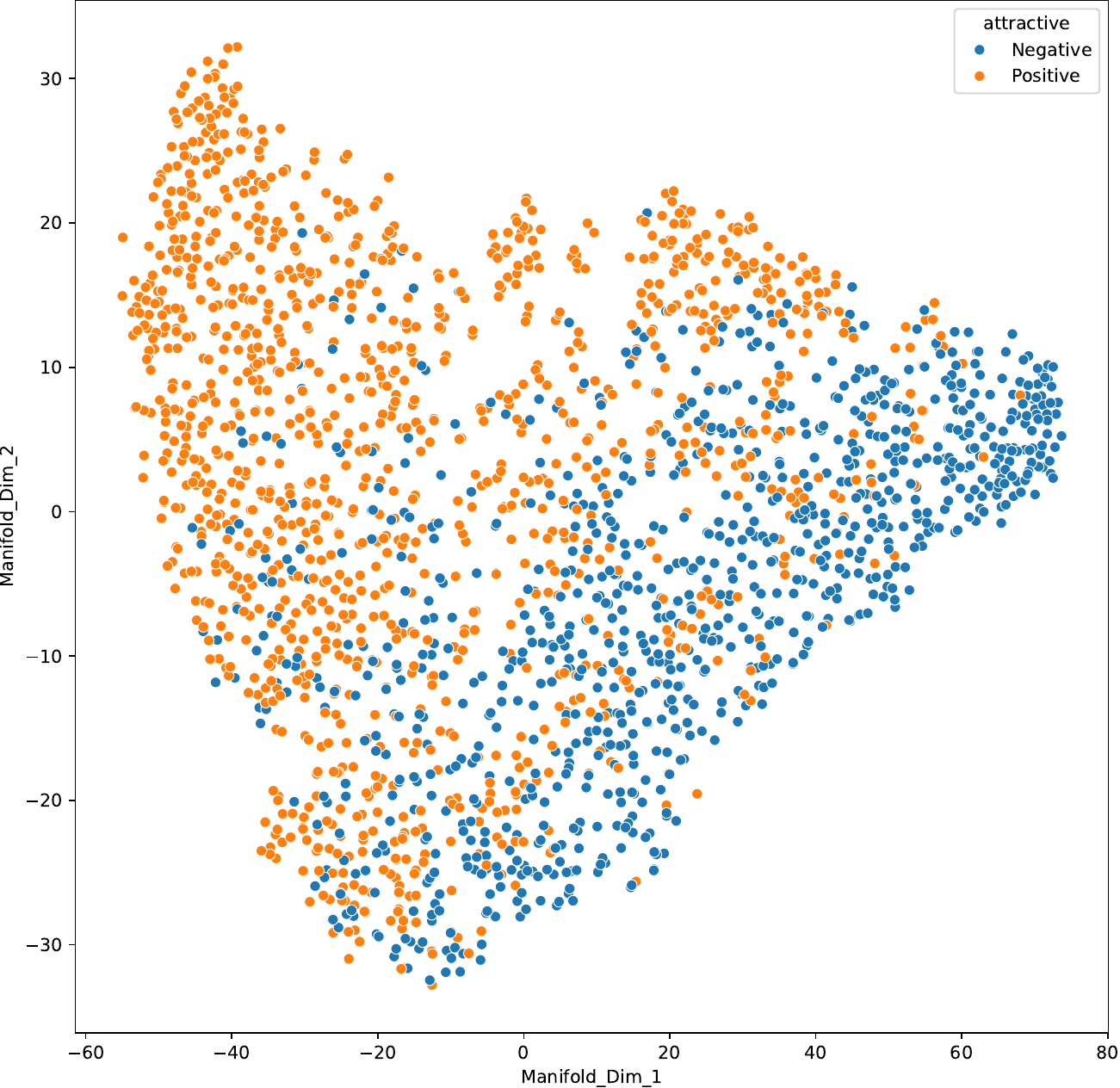}
        \caption{t-SNE for attractive}
    \end{subfigure}
    \begin{subfigure}{0.4\textwidth}
        \includegraphics[width=1.0\linewidth]{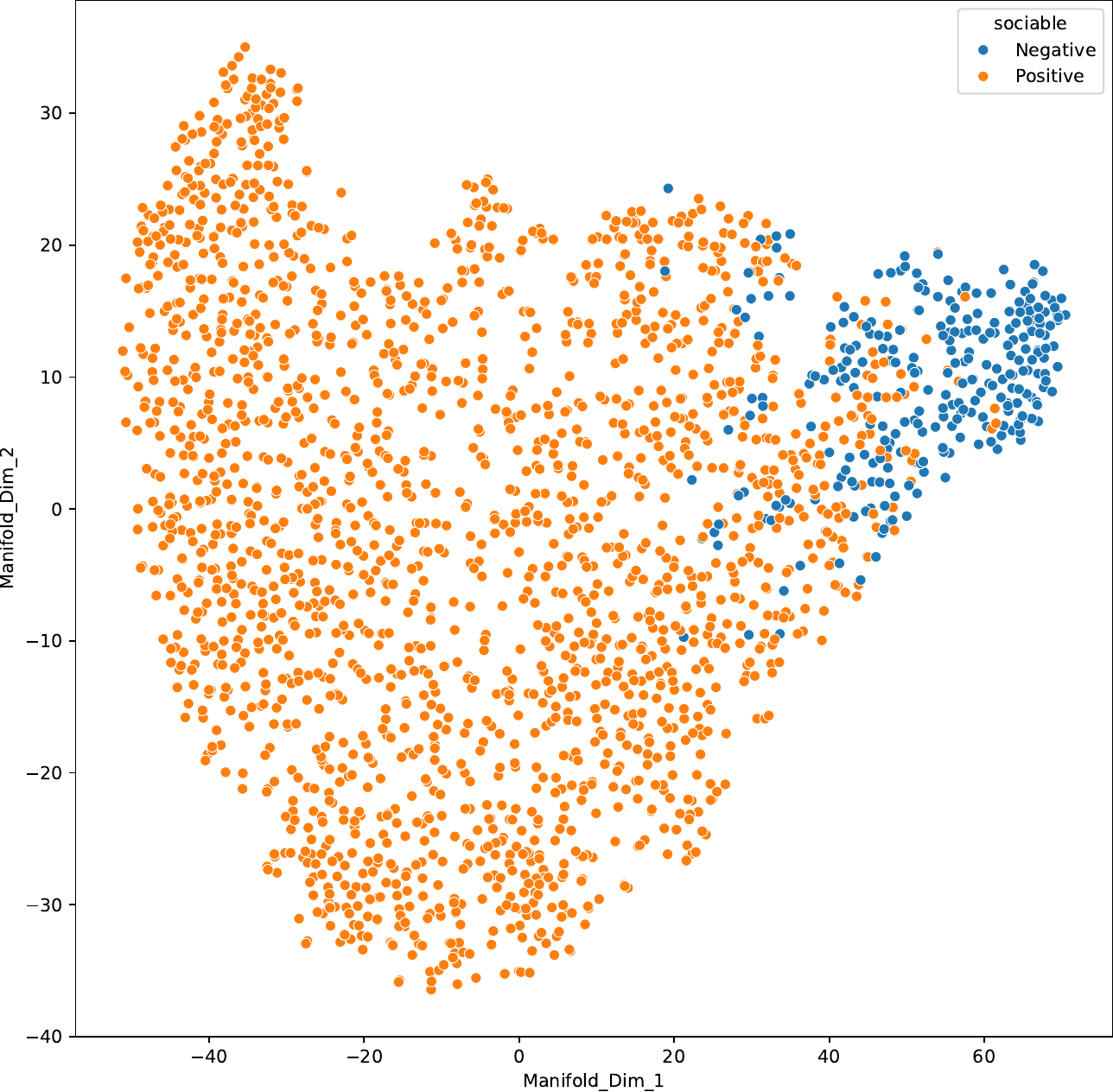}
        \caption{t-SNE for sociable}
    \end{subfigure}
    \begin{subfigure}{0.4\textwidth}
        \includegraphics[width=1.0\linewidth]{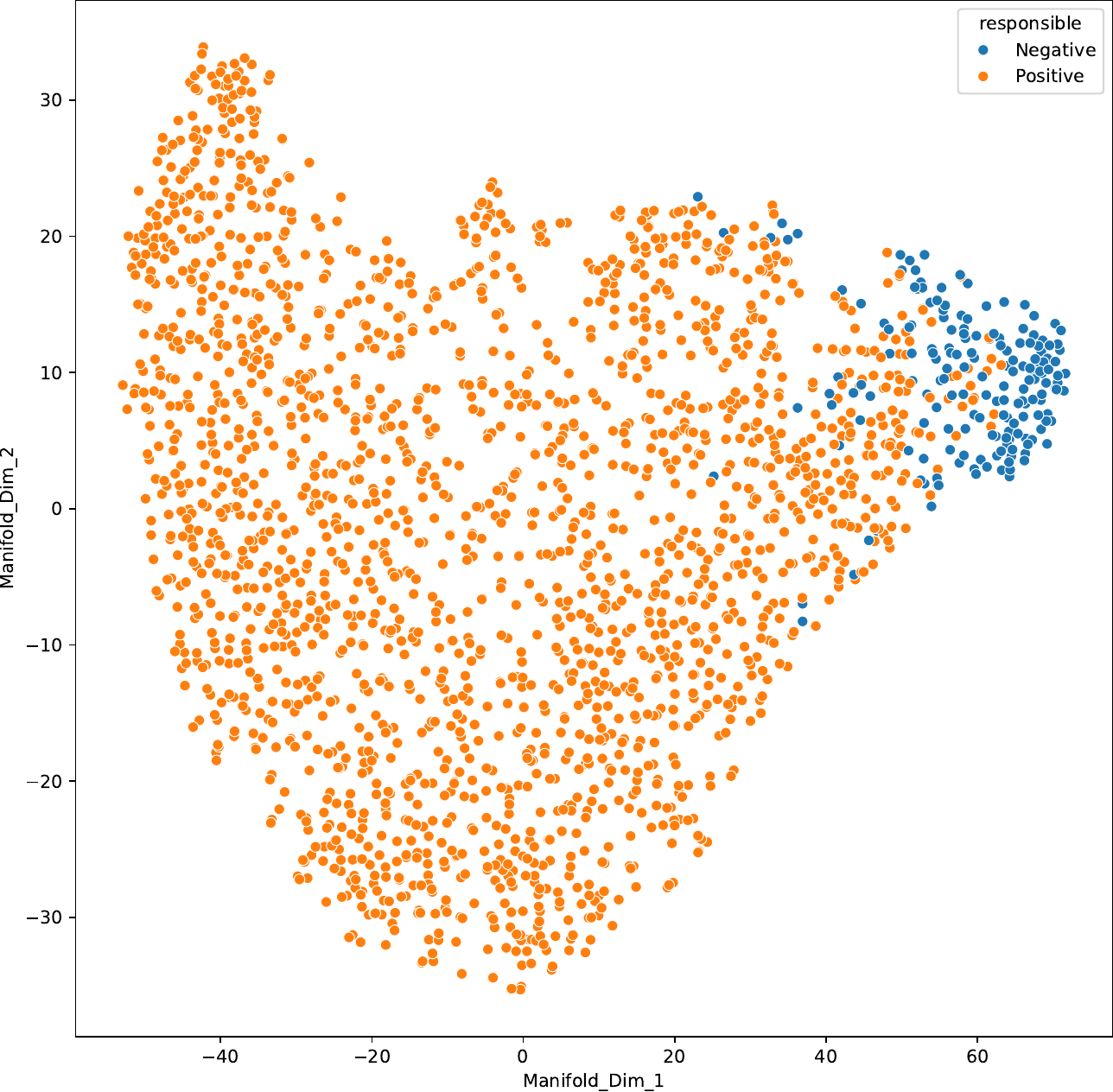}
        \caption{t-SNE for responsible}
    \end{subfigure}
    \caption{t-SNE manifold visualization}\label{fig:manifold_tsne}
\end{figure*}

\begin{figure*}
    \centering
    \begin{subfigure}{0.4\textwidth}
        \includegraphics[width=1.0\linewidth]{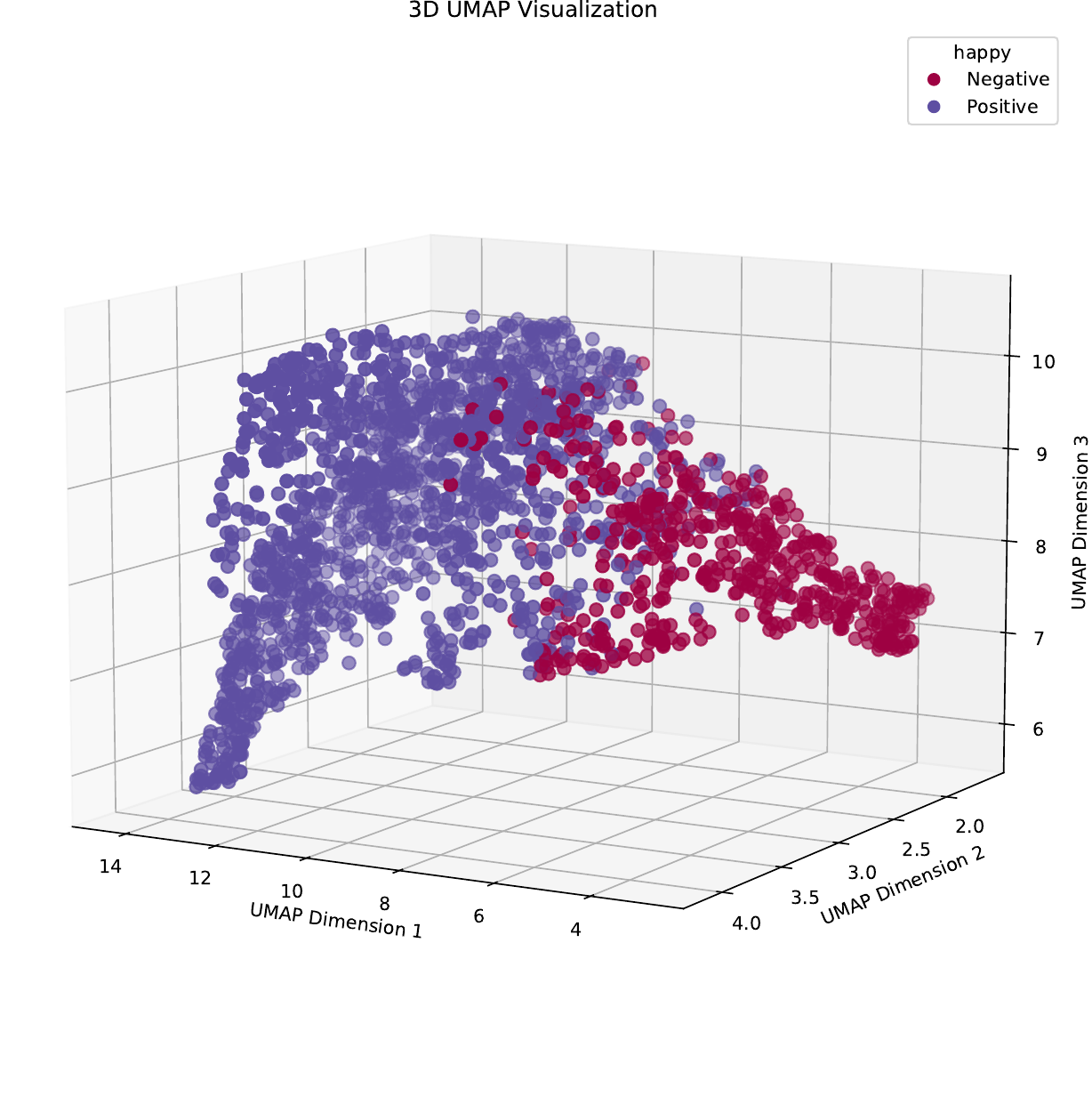}
        \caption{UMAP for happy}
    \end{subfigure}
    \begin{subfigure}{0.4\textwidth}
        \includegraphics[width=1.0\linewidth]{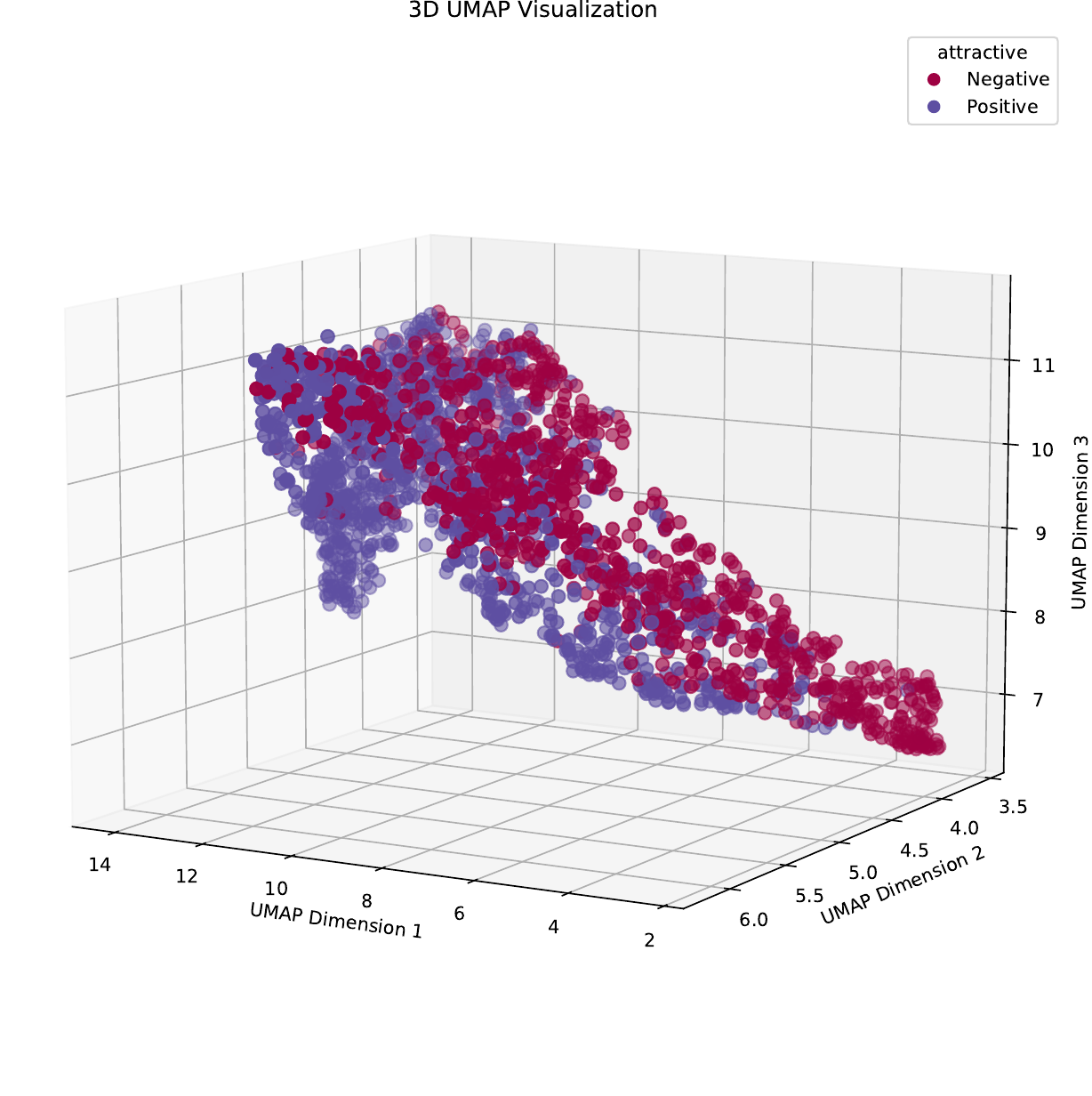}
        \caption{UMAP for attractive}
    \end{subfigure}
    \begin{subfigure}{0.4\textwidth}
        \includegraphics[width=1.0\linewidth]{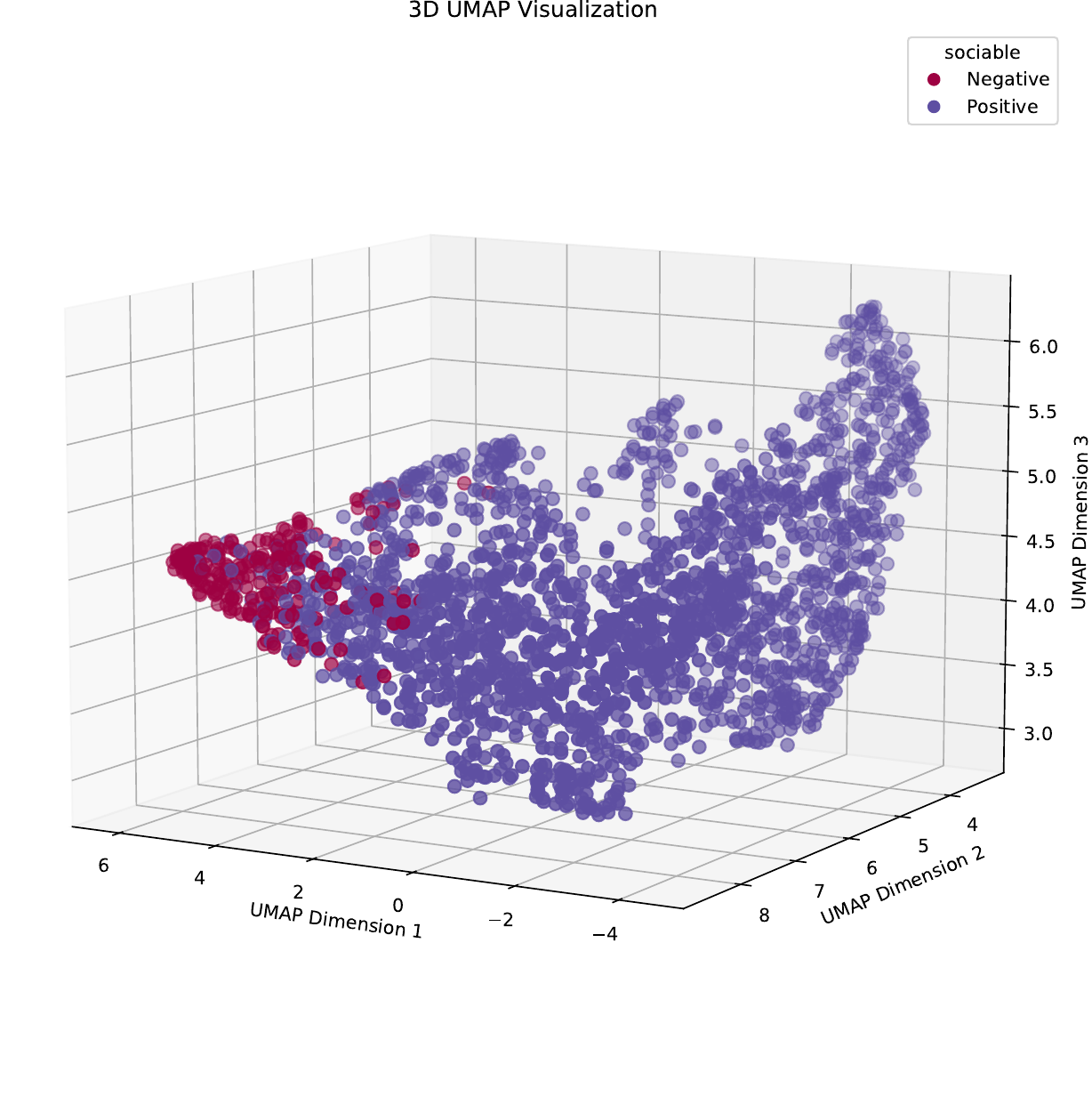}
        \caption{UMAP for sociable}
    \end{subfigure}
    \begin{subfigure}{0.4\textwidth}
        \includegraphics[width=1.0\linewidth]{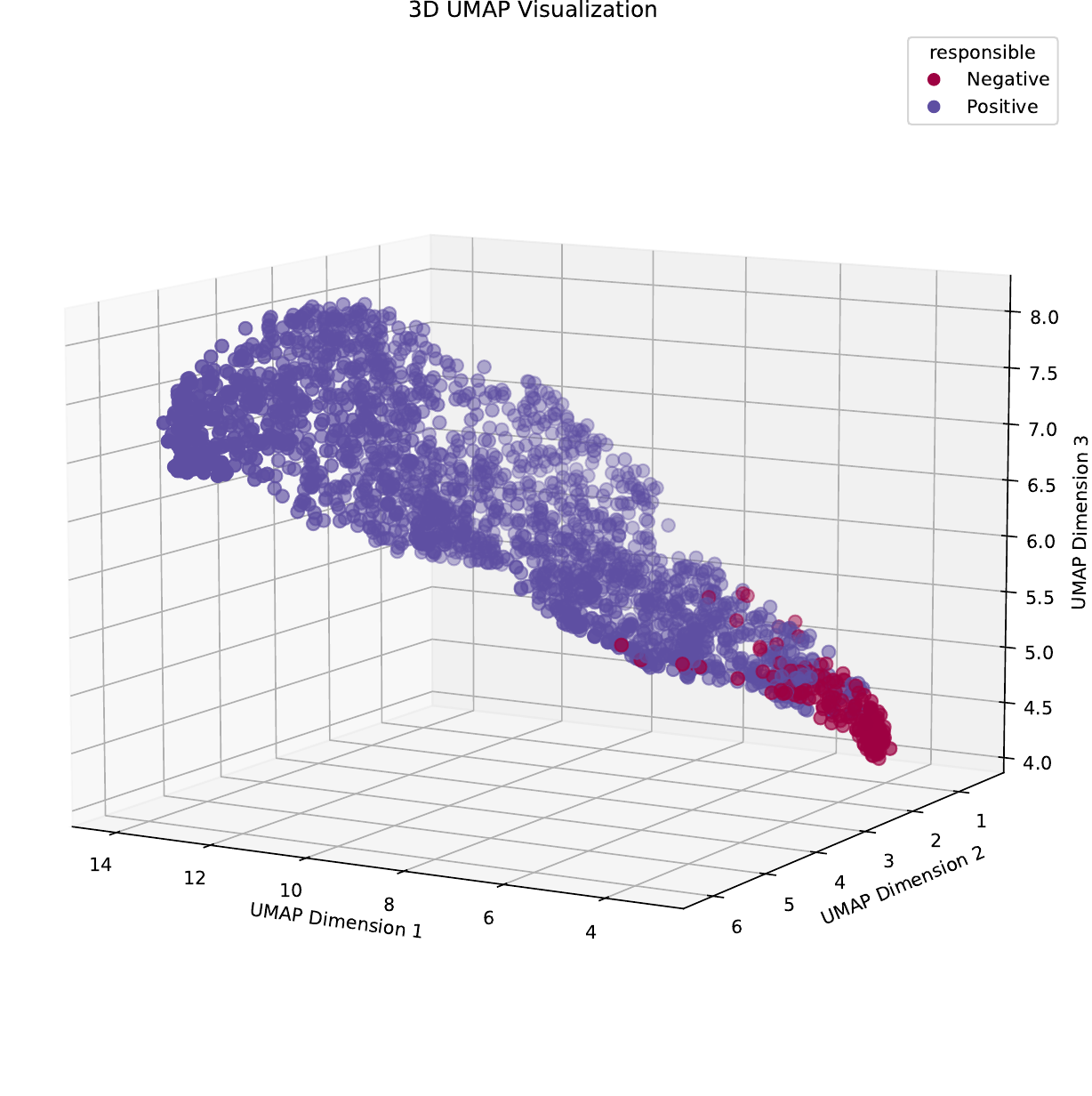}
        \caption{UMAP for responsible}
    \end{subfigure}
    \caption{UMAP manifold visualization}\label{fig:manifold_umap}
\end{figure*}

We use t-SNE and UMAP to visulize the manifold explanation of psychological attributes from 2D and 3D perspectives. Results are presented in fig. \ref{fig:manifold_tsne} and \ref{fig:manifold_umap}. Taking these four psychological attributes as an example, apart from attractive, manifold explanations show the distinct clusters that could reflect inherent categories or patterns within the dataset. 

\newpage
\section{Discussion}
\subsection{Objective cognition or subjective feeling?}
From a cognitive standpoint, attractiveness can be quantified using specific facial or bodily features that align with symmetrical, average, and other evolutionarily favorable traits\cite{fan_bottom-up_2020}. Research using deep learning has shown that machines can predict perceived attractiveness with significant accuracy when trained on large datasets annotated with attractiveness ratings. These models, such as the VGG16 network utilized in our experiments, identify and weigh visual features that correlate strongly with human judgments of attractiveness.

However, some psychological attributes (such as attractiveness) do not form clear clusters in manifold explanations (see fig \ref{fig:manifold_tsne} and fig \ref{fig:manifold_umap}).

This implies that there are additional variables to consider incorporating into the high-dimensional space of psychological data. Merely using facial image data and 20 pairs of psychological attributes is insufficient for a comprehensive explanation of certain psychological traits.

\begin{figure*}[ht]
    \centering
    \begin{minipage}{0.45\textwidth}
        \centering
        \includegraphics[width=\linewidth]{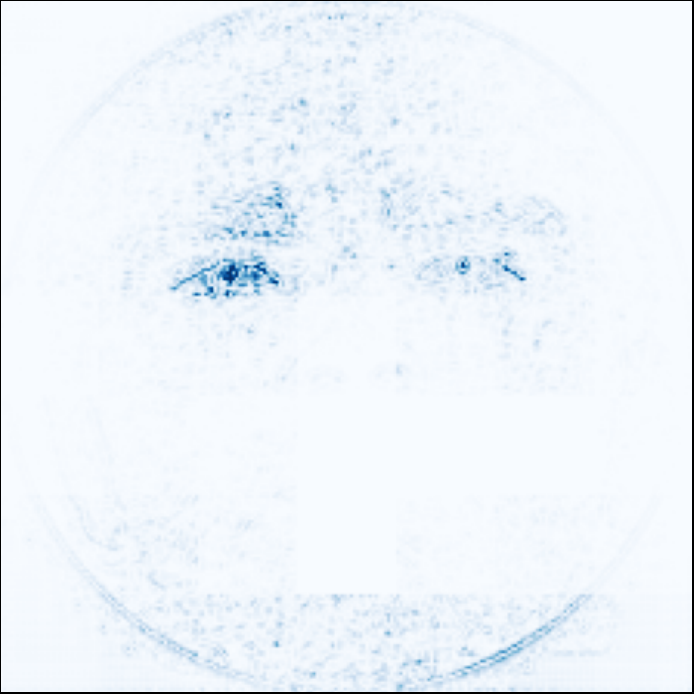}
    \end{minipage}
    \begin{minipage}{0.45\textwidth}
        \centering
        \includegraphics[width=\linewidth]{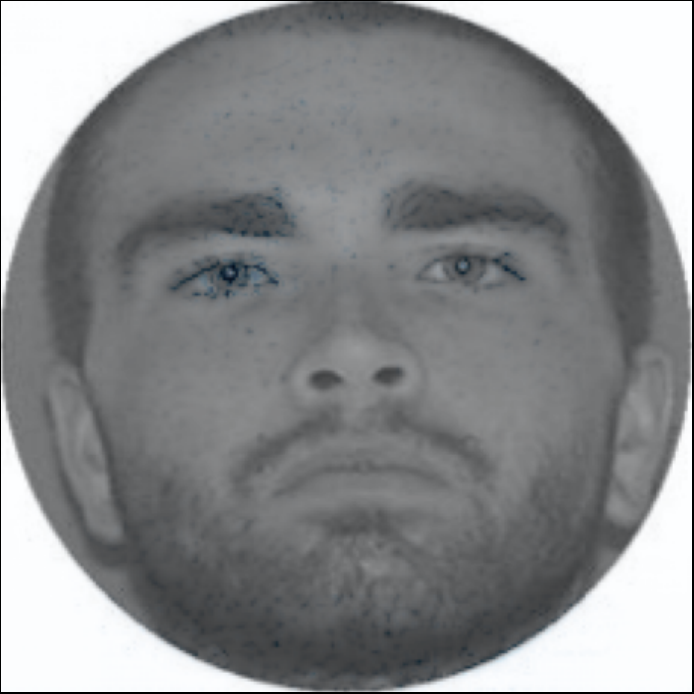}
    \end{minipage}
    \begin{minipage}{0.45\textwidth}
        \centering
        \includegraphics[width=\linewidth]{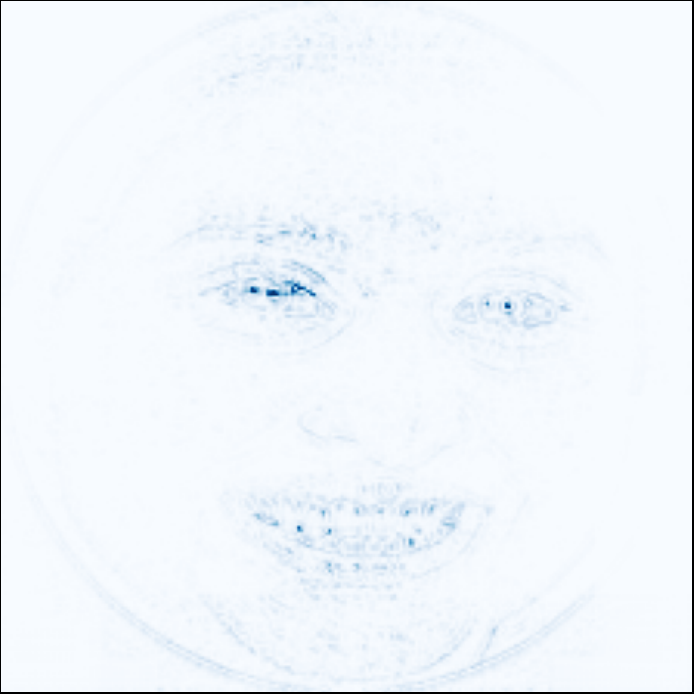}
    \end{minipage}
    \begin{minipage}{0.45\textwidth}
        \centering
        \includegraphics[width=\linewidth]{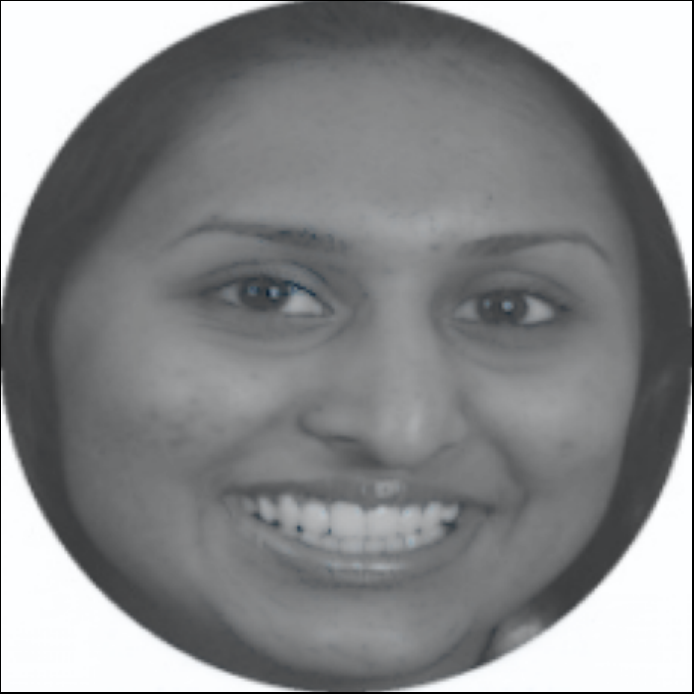}
    \end{minipage}
    \caption{Target: attractive}\label{fig:attractive_post-hoc}
\end{figure*}

Actually, Face preferences affect a diverse range of critical social outcomes, from mate choices and decisions about platonic relationships to hiring decisions and decisions about social exchange\cite{little2011facial}. There are many important sources of individual differences in face preferences (e.g. hormone levels and fertility, own attractiveness and personality, visual experience, familiarity and imprinting, social learning). 

From the manifold explanation learned by neural networks, it can be seen that psychological attributes such as happy, responsible, and sociable exhibit obvious clustering. However, for the attribute of attractive, there is no clear boundary in both two-dimensional and three-dimensional manifold explanations, indicating that neural networks have not fully grasped this rule under existing variables.

\subsection{Ethics bias visualization through manifold}
Humans have a clear personal preference for annotating facial expression images. We have demonstrated through manifold technology that neural networks also exhibit such decision biases. This suggests that, despite their computational nature, neural networks may inherently learn and replicate human-like biases when trained on datasets created by human annotators. The implications of these findings are significant, especially considering the potential for AI systems to inherit and perpetuate societal biases present in the training data.
\section{Conclusion}
In summary, we have achieved SOTA for predictive power and discussed the reasons for neural network decision-making and potential biases from the perspectives of post-hoc and manifold explanations
\appendix
\begin{table}[ht]
    \centering
    \begin{tabular}{lcccc}
    \hline
    Attributes & Human & Geometric & VGG16 & Our Method \\ \hline
    Happy & 0.84 & 0.86 & 0.84 & \textbf{0.91} \\
    Unhappy & 0.75 & 0.81 & 0.8 & \textbf{0.87} \\
    Friendly & 0.78 & 0.83 & 0.82 & \textbf{0.88} \\
    Unfriendly & 0.72 & 0.8 & 0.79 & \textbf{0.83} \\
    Sociable & 0.74 & 0.78 & 0.78 & \textbf{0.84} \\
    Introverted & 0.5 & 0.64 & 0.65 & \textbf{0.7} \\
    Attractive & 0.72 & 0.66 & 0.75 & \textbf{0.8} \\
    Unattractive & 0.62 & 0.62 & 0.7 & \textbf{0.75} \\
    Kind & 0.72 & 0.79 & 0.79 & \textbf{0.83} \\
    Mean & 0.69 & 0.75 & 0.73 & \textbf{0.8} \\
    Caring & 0.72 & 0.78 & 0.79 & \textbf{0.83} \\
    Cold & 0.71 & 0.81 & 0.79 & \textbf{0.85} \\
    Trustworthy & 0.62 & 0.72 & 0.73 & \textbf{0.78} \\
    Untrustworthy & 0.6 & 0.69 & 0.7 & \textbf{0.75} \\
    Responsible & 0.58 & 0.65 & 0.7 & \textbf{0.73} \\
    Irresponsible & 0.55 & 0.67 & 0.67 & \textbf{0.7} \\
    Confident & 0.55 & 0.55 & 0.61 & \textbf{0.65} \\
    Uncertain & 0.45 & 0.62 & 0.63 & \textbf{0.65} \\
    Humble & 0.55 & 0.64 & 0.63 & \textbf{0.65} \\
    Egotistic & 0.52 & 0.62 & 0.62 & \textbf{0.64} \\
    Emotionally Stable & 0.53 & 0.64 & 0.67 & \textbf{0.71} \\
    Emotionally Unstable & 0.5 & 0.62 & 0.64 & \textbf{0.66} \\
    Normal & 0.49 & 0.58 & 0.61 & \textbf{0.67} \\
    Weird & 0.52 & 0.5 & 0.56 & \textbf{0.58} \\
    Intelligent & 0.49 & 0.53 & \textbf{0.62} & \textbf{0.62} \\
    Unintelligent & 0.43 & 0.53 & 0.62 & \textbf{0.63} \\
    Interesting & 0.42 & 0.64 & \textbf{0.67} & \textbf{0.67} \\
    Boring & 0.39 & 0.54 & 0.6 & \textbf{0.61} \\
    Calm & 0.41 & 0.47 & \textbf{0.50} & \textbf{0.5} \\
    Aggressive & 0.65 & 0.72 & 0.72 & \textbf{0.74} \\
    Emotional & 0.33 & 0.6 & 0.6 & \textbf{0.63} \\
    Unemotional & 0.56 & 0.76 & 0.75 & \textbf{0.78} \\
    Memorable & 0.3 & 0.38 & 0.48 & \textbf{0.56} \\
    Forgettable & 0.27 & 0.4 & 0.48 & \textbf{0.51} \\
    Typical & 0.28 & 0.41 & \textbf{0.43} & \textbf{0.43} \\
    Atypical & 0.24 & 0.4 & \textbf{0.43} & \textbf{0.43} \\
    Common & 0.25 & 0.37 & 0.4 & \textbf{0.42} \\
    Uncommon & 0.27 & 0.38 & 0.4 & \textbf{0.41} \\
    Familiar & 0.24 & 0.42 & 0.44 & \textbf{0.49} \\
    Unfamiliar & 0.18 & 0.4 & 0.44 & \textbf{0.45} \\
    \hline
    \end{tabular}
    \caption{Predictive Power (Pearson Correlation Coefficients)}\label{tab:results}
\end{table}
\bibliographystyle{splncs04} 
\bibliography{gra}

\end{document}